\documentclass[]{spie}

\usepackage{float}
\usepackage{tabularx}
\usepackage{graphicx,graphics}
\usepackage{url}
\usepackage{amsmath,amsfonts,amssymb,textcomp,booktabs,threeparttable}
\usepackage{amssymb}
\usepackage{hhline}
\usepackage{multirow}
\usepackage[bookmarks=true,bookmarksopen=true,breaklinks=true,colorlinks=true, linkcolor=blue,anchorcolor=blue,citecolor=blue,filecolor=blue,menucolor=blue, urlcolor=blue,bookmarks=false]{hyperref}
\usepackage{stmaryrd}
\include{boldcommands}
\usepackage{tabularx}
\usepackage{booktabs}
\usepackage{subcaption}
\usepackage{eucal,epstopdf}

\title{Extracting 2D weak labels from volume labels using multiple instance learning in CT hemorrhage detection}

\author[1,2,3,4]{Samuel W. Remedios}
\author[4]{Zihao Wu}
\author[5]{Camilo Bermudez}
\author[4]{Cailey I. Kerley}
\author[1]{Snehashis Roy}
\author[7]{Mayur B. Patel}
\author[2]{John A. Butman}
\author[4,5,6]{Bennett A. Landman}
\author[1,2]{Dzung L. Pham}

\affil[1]{Center for Neuroscience and Regenerative Medicine, Henry Jackson Foundation}
\affil[2]{Radiology and Imaging Sciences, Clinical Center, National Institute of Health}
\affil[3]{Department of Computer Science, Middle Tennessee State University}
\affil[4]{Department of Electrical Engineering, Vanderbilt University}
\affil[5]{Department of Biomedical Engineering, Vanderbilt University}
\affil[6]{Department of Computer Science, Vanderbilt University}
\affil[7]{Departments of Surgery, Neurosurgery, Hearing \& Speech Sciences; Center for Health Services Research, Vanderbilt Brain Institute; Critical Illness, Brain Dysfunction, and Survivorship Center, Vanderbilt University Medical Center; VA Tennessee Valley Healthcare System, Department of Veterans Affairs Medical Center}

\begin{document}
\maketitle

\begin{abstract}
Multiple instance learning (MIL) is a supervised learning methodology that aims to allow models to learn instance class labels from bag class labels, where a bag is defined to contain multiple instances. MIL is gaining traction for learning from weak labels but has not been widely applied to 3D medical imaging.  MIL is well-suited to clinical CT acquisitions since (1) the highly anisotropic voxels hinder application of traditional 3D networks and (2) patch-based networks have limited ability to learn whole volume labels. In this work, we apply MIL with a deep convolutional neural network to identify whether clinical CT head image volumes possess one or more large hemorrhages ($> 20$cm$^3$), resulting in a learned 2D model without the need for 2D slice annotations. Individual image volumes are considered separate bags, and the slices in each volume are instances. Such a framework sets the stage for incorporating information obtained in clinical reports to help train a 2D segmentation approach.  Within this context, we evaluate the data requirements to enable generalization of MIL by varying the amount of training data. Our results show that a training size of at least $400$ patient image volumes was needed to achieve accurate per-slice hemorrhage detection. Over a five-fold cross-validation, the leading model, which made use of the maximum number of training volumes, had an average true positive rate of $98.10\%$, an average true negative rate of $99.36\%$, and an average precision of $0.9698$. The models have been made available along with source code\cite{mil_github} to enabled continued exploration and adaption of MIL in CT neuroimaging. 

\keywords{multiple instance learning, deep learning, neural network, computed tomography (CT), hematoma, lesion, classification}
\end{abstract} 

\section{Introduction}
\label{sec:intro}
\begin{figure*}
    \begin{center}
    \includegraphics[width=1\textwidth]{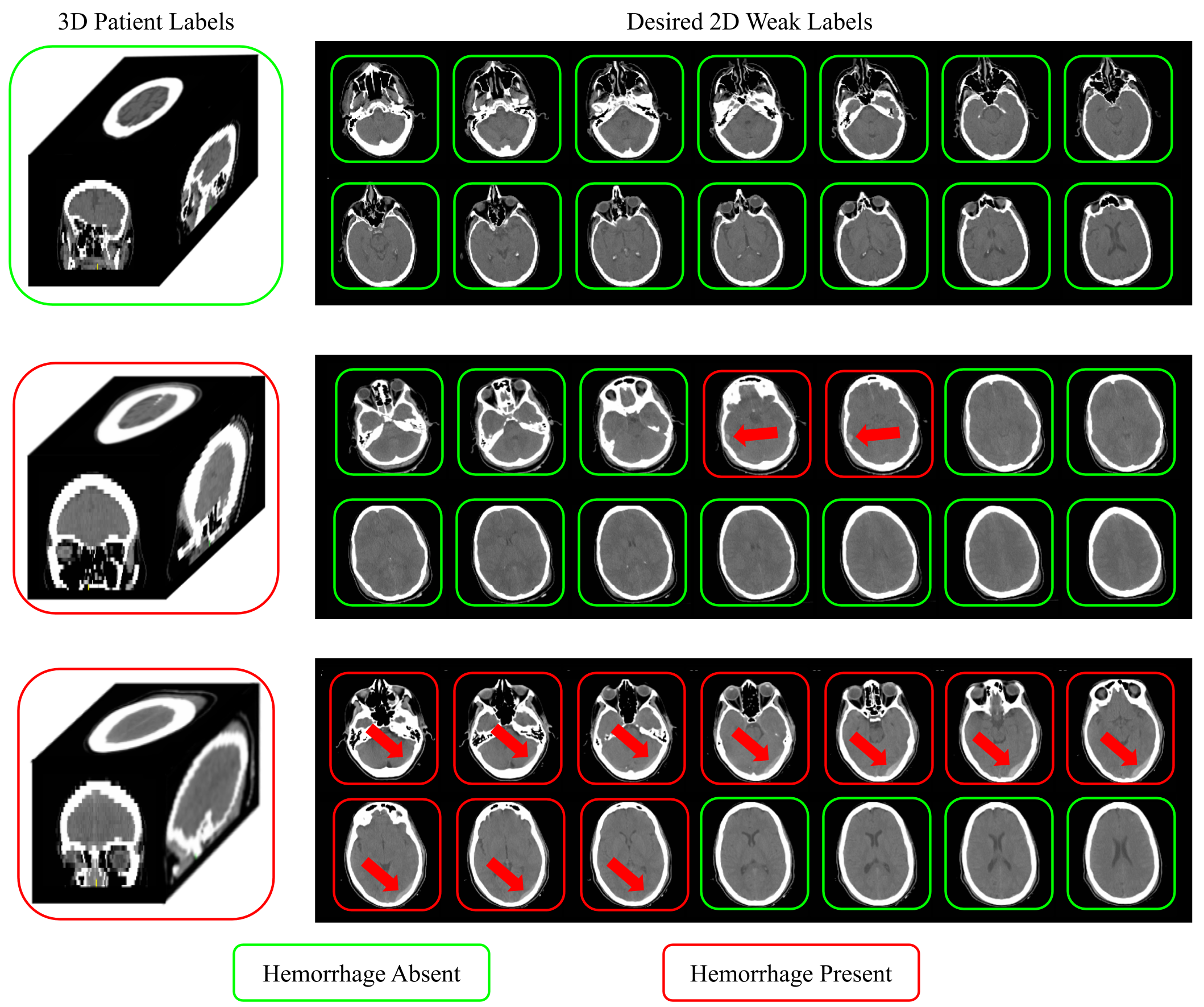}
    \end{center}
    \caption{Radiological diagnoses are weak 3D patient labels, and do not necessarily contain information on the location of that diagnosis.  Weak 2D per slice labels are useful for image analysis and 2D convolutional neural networks. Here, green boxes correspond to negative samples, images with the absence of a hemorrhage.  Red boxes correspond to positive samples, images with the presence of a hemorrhage.  We aim to increase the amount of label information in a 3D volume by extracting 2D weak labels from 3D weak labels.}
    \label{fig:the_problem}
\end{figure*}

Radiological interpretations are commonly available for clinically acquired medical images.  There is growing interest in incorporating such information into medical image analysis algorithms, particularly in the segmentation and quantification of lesions or other pathology within the image.   Segmentation algorithms can be facilitated by knowing whether a lesion is present a priori using “weakly” supervised machine learning approaches\cite{de2016machine}.  However, many segmentation algorithms are slice or patch-based, providing an additional challenge that the clinical determination of the presence of a lesion within the volume may or may not apply to that local slice or patch.  The goal of this work is to learn 2D features accurately from weak 3D volumetric patient labels without the need for 2D manual annotation or interpolation into isotropic 3D space (Figure~\ref{fig:the_problem}). 

Lesions may occur in a number of neurological conditions, including traumatic brain injury (TBI), stroke, and cancer.  Here, we focus on the detection of cerebral hemorrhages due to TBI in computed tomography (CT) scans.  However, the acquisition of CT does not always provide isotropic images, especially in clinical environments where shorter scan durations are desired for safety and patient concerns.  To minimize acquisition time and maximize signal-to-noise ratio, clinical CT scans are frequently acquired with low through-plane resolution, preserving high in-plane resolution (typically on the order of $0.5\times0.5\times5$mm$^3$).  While clinicians can visually interpret such thick-slice images, contemporary deep learning approaches must handle this anisotropy in other ways.

Historically, supervised deep learning applied to 3D medical images either makes use of isotropic research scans or anisotropic clinical scans that must be spatially interpolated.  Traditional interpolation, however, only increases voxel density, not frequency resolution, and partial volume effects interfere with image processing pipelines such as registration.  As such, Zhao et al. addressed anisotropy by applying self-trained super resolution with deep learning, interpolating both spatial and frequency information\cite{zhao2018deep}.  Others addressed anisotropy with network design rather than data preprocessing. Li et al.~\cite{li2018h} first processed 2D slices of a volume before concatenating these slices into a 3D network.  Chen et al.~\cite{chenlstm2d} used a recurrent neural network, effective at sequential data, to handle consecutive 2D slices. Liu et al.~\cite{liu_anisotropic} proposed to use a hybrid network to transfer 2D features to a 3D network.  Lee et al~\cite{lee2017superhuman} selectively skipped feature downsampling in the low-resolution dimension, used 2D features at the finest bottleneck point, and applied anisotropic convolutional filters.  Many others\cite{ciompi2015automatic}$^,$\cite{zhou2016three}$^,$\cite{de20162d} restrict deep learning models to 2D, acting on the high resolution slices as they were acquired.  This approach, however, requires manual annotation on each slice and loses 3D contextual information.

Recently, multiple instance learning (MIL)\cite{dietterich1997solving} has been proposed for use in tandem with deep learning \cite{wu2015deep}$^,$\cite{yan2016multi}.  In MIL, a ``bag" is defined to be a collection of ``instances", and a bag is classified as positive if any instance is positive and is only considered negative if all instances are negative. In the pathology domain, MIL has been used to classify very high resolution slides as cancerous. In this case, a bag was the entire slide and instances were non-overlapping patches extracted from the slide\cite{campanella2018terabyte}.  MIL has also been applied to classify mammograms\cite{zhu2017deep}, detect gastric cancer in abdominal CT\cite{li2015multiple}, detect colon cancer in abdominal CT\cite{dundar2008multiple}$^,$\cite{wu2009min}, classify nodules in chest CT\cite{safta2018multiple}, and classify chronic obstructive pulmonary disease in chest CT\cite{cheplygina2014classification}.  Particularly of note, MIL has been used alongside deep learning to detect malignant cancerous nodules in lung CT\cite{shen2016learning} and to segment cancer in histopathology\cite{jia2017constrained}.

For segmentation of CT hemorrhages, 2D convolutional neural networks (CNNs) have demonstrated good success.  To augment such a model with weak labels indicating presence of a hemorrhage within the volume, we consider a bag to be the collection of 2D slices within the volume, and an instance to be the 2D slice.  We investigate the potential of using deep learning to predict whether a hemorrhage is present in  a slice using MIL.  We first train a 2D CNN to classify the presence of hemorrhages.  Then, to characterize the minimum number of image volumes required to train such a model, we vary the number of training samples from $100$ to $672$, the maximum available number of training samples in our dataset.  To the best of our knowledge this is the first application of MIL to extract 2D features from 3D labels. Furthermore, this is the first characterization of the number of bags necessary for a MIL task using deep neural networks.

\section{Method}
\label{sec:method}

Herein we describe the format and preprocessing of the data as well as the implementation of MIL.  MIL allows models to learn a binary classification task by learning from one or some instances in the bag.  We selected ``max pooling" as our MIL pooling algorithm.  ``Max pooling" is an overloaded term in machine learning; in the MIL context, ``max pooling" refers to the selection of the ``most-positive" instance\cite{wang2019comparison}.  On the surface it may not be immediately apparent how a neural network trained with MIL is able to achieve convergence with a max pooling operation.  Given a randomly initialized neural network, the ``most-positive" instance on the forward pass is not guaranteed to be a truly positive instance.  The main mechanism which allows MIL neural networks to converge is the setup of the training data: a bag which is negative is guaranteed to contain all negative instances.  In this way, the model learns which instances are negative, and positive instances (which in our case present differently in the images) are anomalies, and thus the model can learn to differentiate the two.

\newpage

\subsection{Data}
\label{subsec:data}
At Vanderbilt University Medical Center, $11,477$ CT image volumes from $4,033$ patients from a consecutive retrospective study of trauma were retrieved in de-identified form under IRB supervision.  All volumes were segmented by a hemorrhage segmentation convolutional neural network (CNN) trained as described in previous work\cite{remedios2019distributed}.  Then, these automatic segmentations were manually reviewed and scored by a trained rater: ($0$) no hemorrhage and correct empty mask, ($1$) hemorrhage and slight errors in mask, ($2$) hemorrhage but large failure of algorithm, ($3$) hemorrhage and near-perfect mask, and ($4$) invalid data for the CNN. Image volumes with scores of $2$ and $4$ were omitted from selection due to uncertainty in hemorrhage size estimation and invalidity, respectively.  All volumes with scores ($0$) were selected as the negative class with no hemorrhage present. From all axially acquired volumes with scores ($1$) and ($3$), image volumes with $>20$ cubic centimeters of blood were selected as the positive class with the presence of severe hemorrhages.  After organizing data as such, the dataset consisted of a total of $4882$ volumes of which $525$ contained hemorrhages and $4357$ did not (Table~\ref{tab:data_distribution}). Since our focus is a hemorrhage detection task, the ground truth labels are a binary classification of the entire image; the aforementioned automatic segmentations were solely used to obtain this binary label and were not used in training any CNN.

\begin{table}[tb]
\caption{Distribution of data for training, validation, and testing.}
\label{tab:data_distribution}
\begin{center}
\begin{tabular}{llll}
\toprule[2pt]
 & Total & Positive Samples & Negative Samples\\
\toprule[1pt]
Train & $672$ & $336$ & $336$\\ 
\cmidrule[1pt]{2-4}
Validation & $168$ & $84$ & $84$\\
\cmidrule[1pt]{2-4}
Test & $4042$ & $105$ & $3937$\\
\toprule[1pt]
Total & $\mathbf{4882}$ & $\mathbf{525}$ & $\mathbf{4357}$\\ 
\bottomrule[2pt]
\end{tabular}
\end{center}
\vspace{-1em}
\end{table}

From the dataset, $20$\% of positive samples were randomly withheld for testing, and of the remaining $80$\%, another $20$\% was randomly selected to be the validation set.  Patients were not mixed in the data split, and the negative samples were randomly undersampled such that classes were balanced for training and validation.  All remaining negative samples were included in the test set.  The sizes of training, validation, and test datasets are shown in Table~\ref{tab:data_distribution}.

All CT image volumes were converted from DICOM to NIFTI using \texttt{dcm2niix}\cite{li2016first} with voxel intensities preserved in Hounsfield units, and as such no intensity normalization was applied.  Subsequently, each volume was skullstripped with \texttt{CT\_BET}\cite{muschelli2015validated} and rigidly transformed to a common orientation.  All axial slices were extracted from the volume and null-padded or cropped to a size of $512 \times 512$ pixels.  These slices were considered ``instances", and were converted to $16$-bit floats before being shuffled and aggregated as ``bags".  All bags were aggregated and written in the TFRecord file format\cite{TFRecords} to avoid memory constraints.

\subsection{Model Hyperparameters}
\label{subsec:model_hyperparams}
The model architecture used for this task was a ResNet-$34$\cite{resnetHe}, with two output neurons activated by \texttt{softmax}.  We trained with a batch size of $128$ facilitated by accumulating gradients.  The learning rate was set to $1 \times 10^{-4}$ with the Adam optimizer.  The loss function was categorical crossentropy.  Convergence was defined as no improvement of validation loss by $1 \times 10^{-4}$ in $50$ epochs.  The selected deep learning framework was Tensorflow\cite{abadi2016tensorflow} v.$1.14$ and an NVIDIA $1080$ TI was used for hardware acceleration.

\subsection{Multiple Instance Learning Implementation}
\begin{figure}[tb]
    \begin{center}
    \includegraphics[width=1\textwidth]{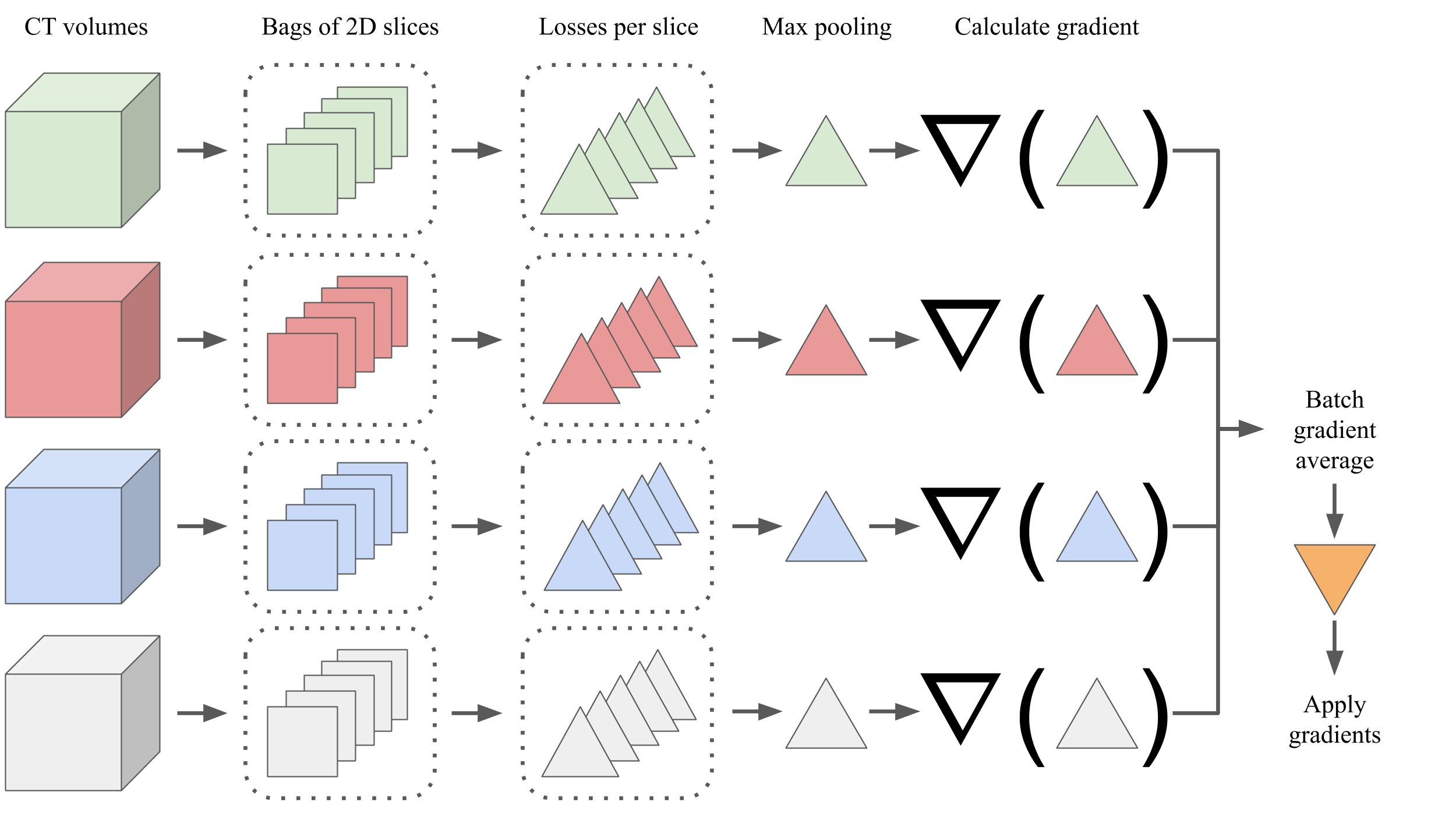}
    \end{center}
    \caption{Illustration of multiple instance learning with gradient accumulation.  The first CT volume is organized into a bag of its 2D slices.  Then, the model performs inference on all slices in the bag, and class probabilities are calculated.  The gradient is calculated only for the instance corresponding to the most probable positive class.  This gradient is saved and the gradient calculation is repeated for the next bag until the batch is done, and the accumulated gradient is averaged for the batch size and applied to the model.}
    \label{fig:mil_diagram}
\end{figure}

Our learning implementation consisted of several steps. First, for each bag, we performed model inference on each instance to obtain the probability that an instance belongs to the hemorrhage positive class.  Then, we identified the instance with the highest probability of being the positive class and calculated the gradients for only this instance.  Each subsequent bag's gradients were aggregated with a running average.  After $128$ bags, the accumulated gradient was applied as a batch to the model.  This process is illustrated in Figure~\ref{fig:mil_diagram}.  In summary, the input to the CNN is a bag of 2D axial slices and the output is the probability to which class this bag belongs.  The resultant model is a 2D CNN which classifies whether an axial CT slice contains a hemorrhage.  Our implementation is publicly available\cite{mil_github}.

\subsection{Dataset Size Restriction}
To investigate the required number of training samples needed for MIL to converge, we trained multiple models from the same initialization point with varying dataset sizes.  We trained a total of $6$ models with datasets comprised of $672$, $500$, $400$, $300$, $200$, and $100$ image volumes randomly selected from the training data outlined in Section~\ref{subsec:data}.  Hereafter, models are referred to as ``Model N", where $N=$ the number of training samples.  All models trained with the same architecture, learning rate, optimizer, and convergence criteria as described in Section~\ref{subsec:model_hyperparams}.

\section{Results}
\begin{figure}
\minipage{0.45\textwidth}
\begin{center}

  \includegraphics[width=0.8\linewidth]{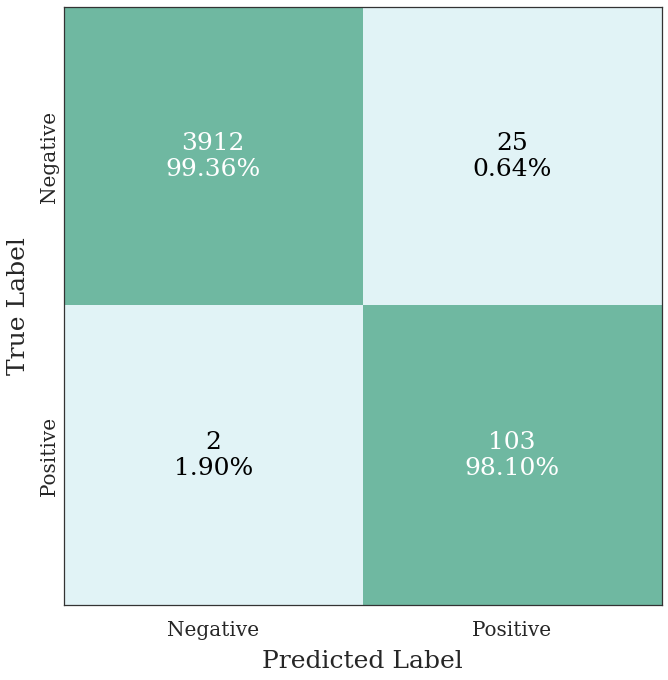}
  \caption{Confusion matrix for Model $672$ obtained after bag-wise classification.  Reported results are an average of five-fold cross-validation. For the negative class, Model $672$ achieved $99.36$\% accuracy with $25$ erroneous classifications, and for the positive class the model achieved $98.10$\% accuracy with $2$ erroneous classifications.}\label{fig:cm}

\end{center}
\endminipage\hfill
\minipage{0.45\textwidth}
\begin{center}

  \includegraphics[width=0.8\linewidth]{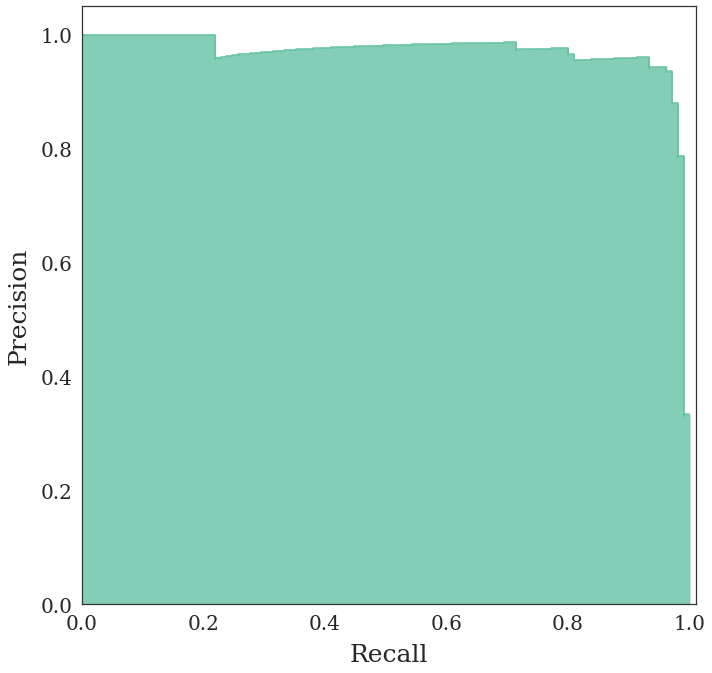}
  \caption{Cross-validated precision-recall curve for Model $672$, trained on the entire training dataset.  The average precision over all folds is $0.9698$.\newline\newline\newline\newline}\label{fig:pr_curve}
  
\end{center}
\endminipage\hfill
\end{figure}

Because validation data on the presence or absence of hemorrhage was not available for each individual slice, evaluation was performed based on 3D classification accuracy. Image volumes were classified to be positive if any slice was positive and are only negative if all slices were negative.  In other words, a patient was considered to have a hemorrhage if there is a hemorrhage present in any 2D slice of the 3D scan.  Thus, Figures~\ref{fig:cm},~\ref{fig:pr_curve},~\ref{fig:ap_per_training_sample}, and~\ref{fig:pr_per_training_sample} report on head CT volumes.

Our convergence criteria terminated model training once the validation loss showed no improvement of $1\times10^{-4}$ in $50$ epochs.  Thus, models of all training set sizes ``converged," but performance on the test set varied as expected.

\subsection{Classification}
First, we consider averaged cross-validated results from the model trained over the entire available training set, Model $672$. Figure ~\ref{fig:cm} shows the class-wise cross-validated testing accuracy of Model $672$.  Overall, the model correctly classified $4015$ of $4042$ image volumes using the MIL paradigm, corresponding to $99.33$\% accuracy. Of the $3937$ image volumes not presenting with hemorrhage, the MIL model attained $99.36$\% accuracy with $25$ false positives.  From the remaining $105$ true positive volumes, the model made $2$ errors.  The corresponding precision-recall curve is shown in Figure~\ref{fig:pr_curve}; Model $672$ achieved an average precision of $0.9698$ over all folds.  

\begin{figure}
\begin{center}
  \includegraphics[width=0.75\linewidth]{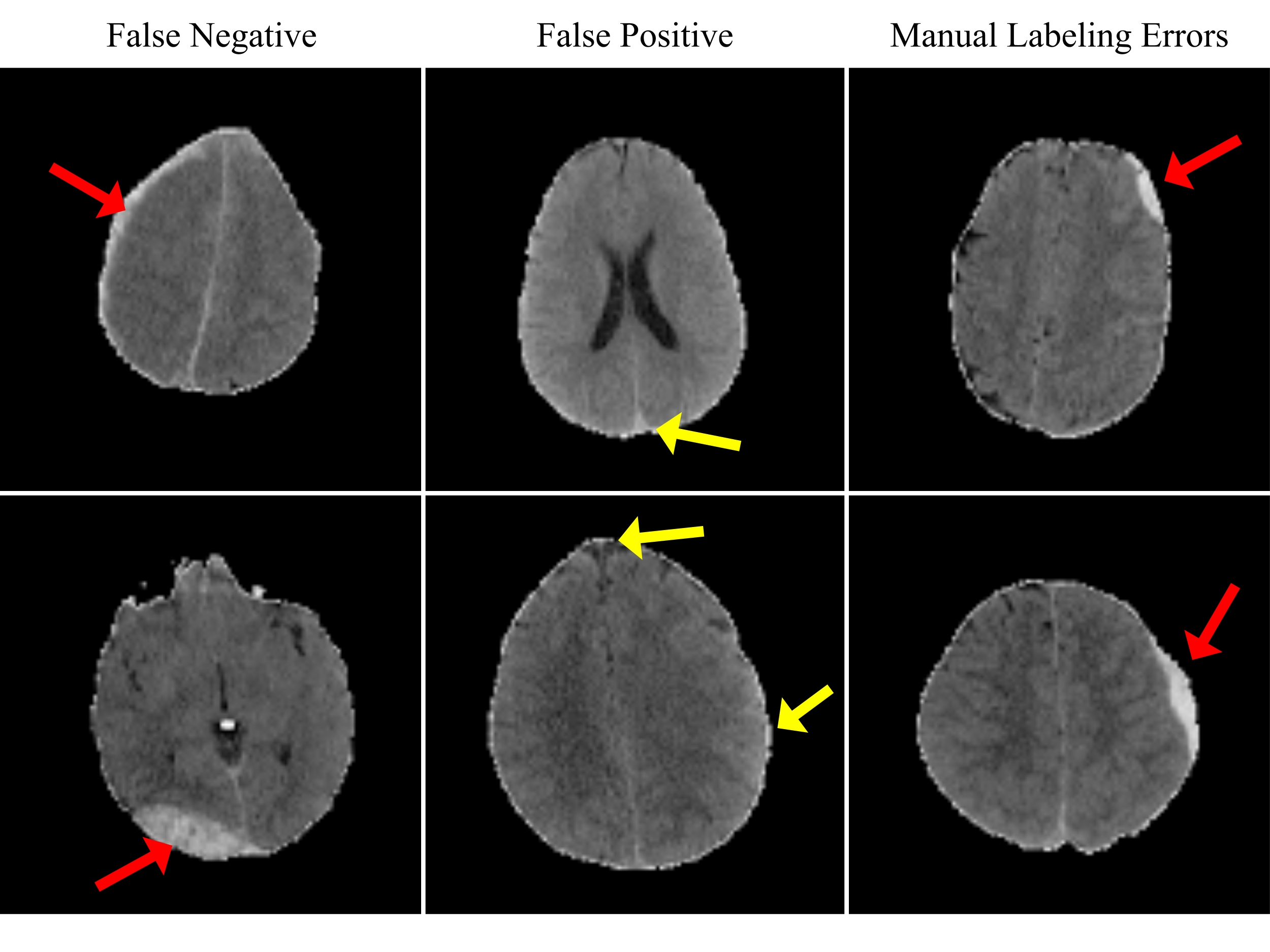}
  \caption{Representative erroneous classifications made by the leading model, Model $672$.  Each displayed axial slice corresponds to the most probable slice containing a hemorrhage. The first column shows false negatives, with apparent hemorrhages indicated by red arrows.  The second column shows false positives, and it is possible that the non-hemorrhage regions indicated by yellow arrows caused the model to consider these volumes positive.  The third column shows manual labeling errors which the model correctly classified as containing hemorrhages, indicated by red arrows.}\label{fig:fp_fn}
\end{center}
\end{figure}

Representative false positives and false negatives are displayed in Figure~\ref{fig:fp_fn}. False positives generally occurred due to bright regions at the brain-skull boundary or in sinuses that were not removed in the skull-stripping step.   Note the third column in Figure~\ref{fig:fp_fn}, corresponding to errors in the manual labels; these false positives were actually true positives where the human rater erroneously classified hemorrhage presence.

\subsection{Reduction of training samples}
\begin{figure}
  \begin{center}
  \includegraphics[width=\linewidth]{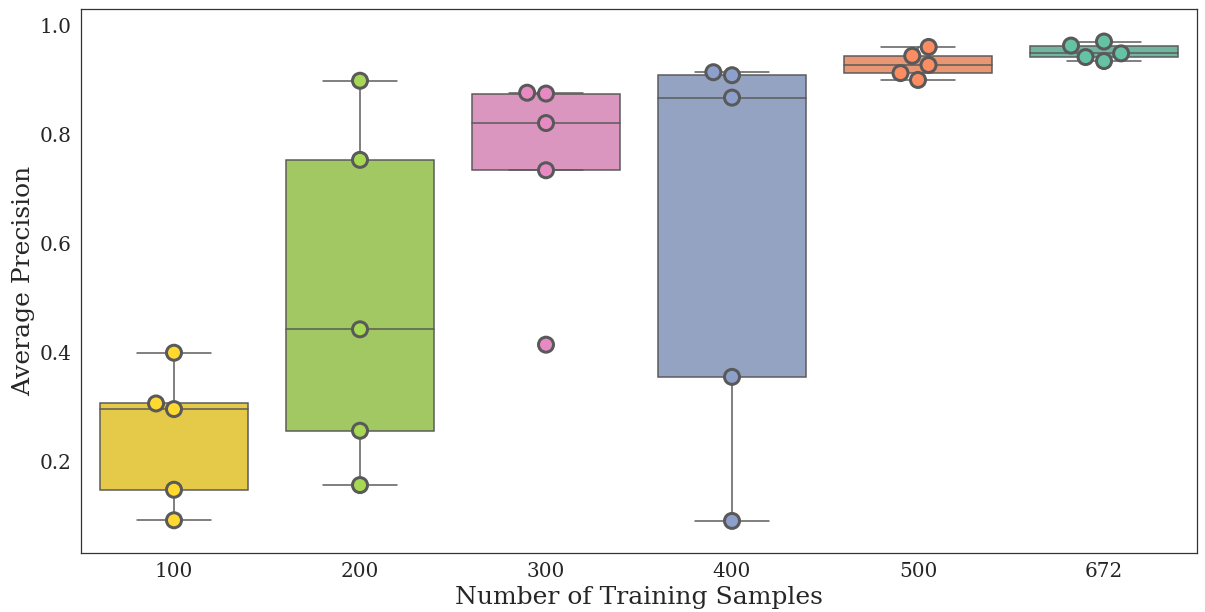}
  \caption{Five-fold cross-validated average precision calculated from precision-recall curves per training sample size.  As expected, average precision increases with more training samples.  Note the inability for the model to learn with MIL at $100$ training samples as well as the model instability with lower numbers of samples.\newline}\label{fig:ap_per_training_sample}
  \end{center}
\end{figure}

\begin{figure}
  \begin{center}
  \includegraphics[width=\linewidth]{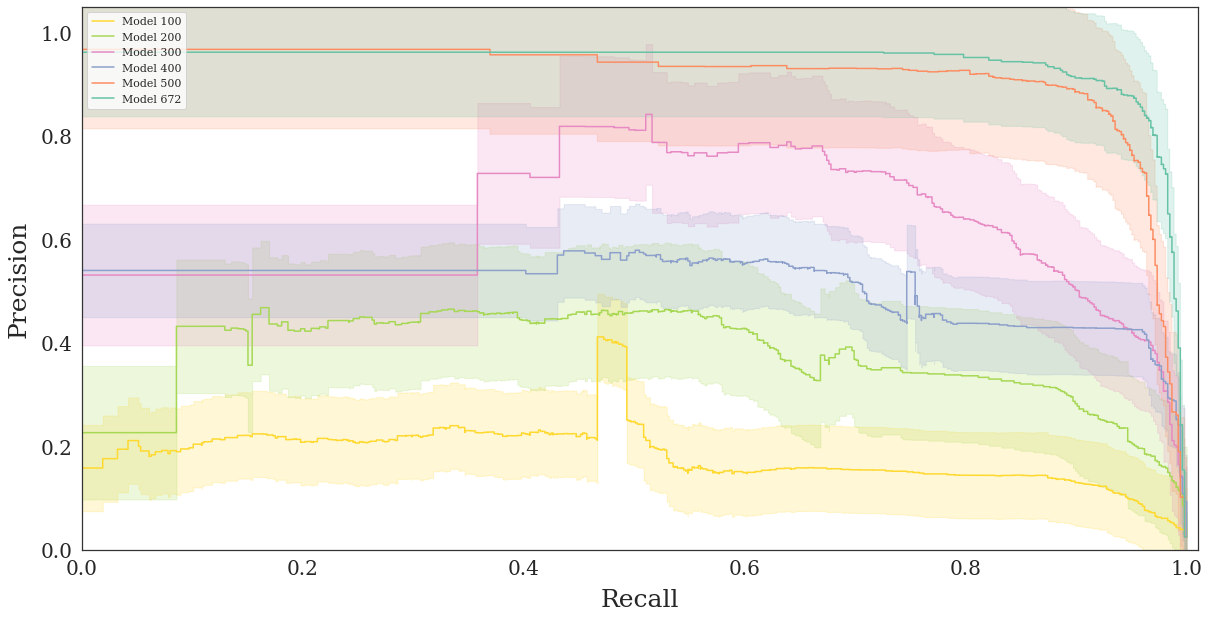}
  \caption{Comparison of precision-recall curves for differing training dataset sizes over five-fold cross-validation.  Faded filled color regions indicate variation in results among folds, and darker lines indicate the mean of results among folds.  Model $100$ was unable to generalize to the testing dataset, where Models $200$ and $300$ achieve moderate performance.  At least $400$ training samples were needed to train a generalizable model with MIL.}\label{fig:pr_per_training_sample}
  \end{center}
\end{figure}

As expected, more data leads to better deep learning model performance, illustrated in Figures~\ref{fig:ap_per_training_sample} and~\ref{fig:pr_per_training_sample}.  Model stability over folds also increases, likely because each fold contains proportionally more data, increasing the likelihood to account for anatomical heterogeneties during training.

A comparison of the average precision obtained by each model trained with varying dataset counts over five folds is shown in Figure~\ref{fig:ap_per_training_sample}.  There were three unexpected results. First, there was a dramatic decrease in model precision when reducing the number of training samples from $200$ to $100$.  While some models trained on folds with $200$ samples performed well, none from folds with $100$ samples achieved the mean performance of models trained with $200$ samples.  Second, there was increased variance of performance in models trained on $400$ compared to those trained on $300$ examples.  Several factors could have influenced this, such as the specific random sampling when constructing the fold or fluctuations in shuffling during the training procedure.  Third, although model performance stabilized across folds with larger number of training samples, the gain in performance from adding more data above $500$ samples was significantly reduced.

\section{Discussion}
The use of 2D supervised CNNs for segmentation of anisotropic medical images is common.  We have applied MIL to allow a 2D CNN to learn from volumetric patient labels on a  hemorrhage detection task. To the best of our knowledge, this is the first application of MIL and deep learning to clinical imaging to circumnavigate the need for 2D image slice labels, as well as the first characterization of required dataset sizes for MIL. We have found that for this hemorrhage detection task, at least $200$ annotated image volumes were necessary for decent classification, and $400$ to train a strong classifier.  We conclude that MIL is a step forward towards building models which learn from patient-level labels.  Moving forward, we anticipate further utility of MIL towards pre-training and transfer learning other weakly supervised tasks. Our source code is publicly available\cite{mil_github}.

\section{ACKNOWLEDGEMENTS}
Support for this work included funding from the Intramural Research Program of the NIH Clinical Center and the Department of Defense in the Center for Neuroscience and Regenerative Medicine, the National Multiple Sclerosis Society RG-1507-05243 (Pham), and NIH grants 1R01EB017230-01A1 (Landman) and R01GM120484 and R01AG058639 (Patel), as well as NSF 1452485 (Landman). The VUMC dataset was obtained from ImageVU, a research resource supported by the VICTR CTSA award (ULTR000445 from NCATS/NIH), Vanderbilt University Medical Center institutional funding and Patient-Centered Outcomes Research Institute (PCORI; contract CDRN-1306-04869). This work received support from the Advanced Computing Center for Research and Education (ACCRE) at the Vanderbilt University, Nashville, TN, as well as in part by ViSE/VICTR VR3029.  We also extend gratitude to NVIDIA for their support by means of the NVIDIA hardware grant.

\bibliographystyle{spiebib}
\small{
\bibliography{remedios-spie2020-refs}
}

\end{document}